\documentclass[letterpaper, 10 pt, journal, twoside]{IEEEtran}  

\ifCLASSINFOpdf
\else
\fi

\usepackage{graphics}
\usepackage{graphicx}
\usepackage{booktabs}
\usepackage{multirow}
\usepackage{amsmath}
\usepackage{amssymb}
\usepackage{booktabs}
\usepackage{float}
\usepackage{balance}
\usepackage{multirow}
\usepackage{pgfplots}

\usepackage[capitalize]{cleveref}
\crefname{section}{Sec.}{Secs.}
\Crefname{section}{Section}{Sections}
\Crefname{table}{Table}{Tables}
\crefname{table}{Tab.}{Tabs.}
\begin{document}
%
\title{You Only Label Once: 3D Box Adaptation from Point Cloud to Image with Semi-Supervised Learning}
%
%
%


\author{Jieqi Shi$^{1}$, Peiliang Li*$^{2}$, Xiaozhi Chen$^{3}$ and Shaojie Shen$^{4}$%
\thanks{Manuscript received: June 27, 2023; Revised August 6, 2023; Accepted August 17, 2023. This paper was recommended for publication by Editor Cesar Cadena upon evaluation of the Associate Editor and Reviewers' comments. This work was supported by The Research Grants Council General Research Fund (RGC GRF) project RMGS20EG20. $^{1}$Jieqi Shi and $^{4}$Shaojie Shen are with Department of Electronic and Computer
Engineering, Hong Kong University of Science and Technology
        {\tt\footnotesize $\{$jshias, eeshaojie$\}$@connect.ust.hk} $^{2}$Peiliang Li and $^{3}$Xiaozhi Chen are with DJI Automotive.
        {\tt\footnotesize $\{$peiliang.uav, cxz.thu$\}$@gmail.com} Digital Object Identifier (DOI): see top of this page.}
}
%
%

\markboth{IEEE Robotics and Automation Letters. Preprint Version. Accepted August, 2023}
{Shi \MakeLowercase{\textit{et al.}}: You Only Label Once: 3D Box Adaptation from Point Cloud to Image with Semi-Supervised Learning} 

%



\maketitle

\begin{abstract}

   The image-based 3D object detection task expects that the predicted 3D bounding box has a ``tightness'' projection (also referred to as cuboid) to facilitate 2D-based training, which fits the object contour well on the image while remaining reasonable on the 3D space. These requirements bring significant challenges to the annotation. Projecting the Lidar-labeled 3D boxes to the image leads to non-trivial misalignment, while directly drawing a cuboid on the image cannot access the original 3D information. In this work, we propose a learning-based 3D box adaptation approach that automatically adjusts minimum parameters of the 360$^{\circ}$ Lidar 3D bounding box to fit the image appearance of panoramic cameras perfectly. With only a few 2D boxes annotation as guidance during the training phase, our network can produce accurate image-level cuboid annotations with 3D properties from Lidar boxes. We call our method ``you only label once'', which means labeling on the point cloud once and automatically adapting to all surrounding cameras. Our refinement balances the accuracy and efficiency well and dramatically reduces the labeling effort for accurate cuboid annotation. Extensive experiments on the public Waymo and NuScenes datasets show that our method can produce human-level cuboid annotation on the image without manual adjustment and can accelerate monocular-3D training tasks.
\end{abstract}

\section{Introduction}
\label{sec:intro}
    \begin{figure}[h]
      \centering
      \fbox{
      \includegraphics[width=0.92\linewidth]{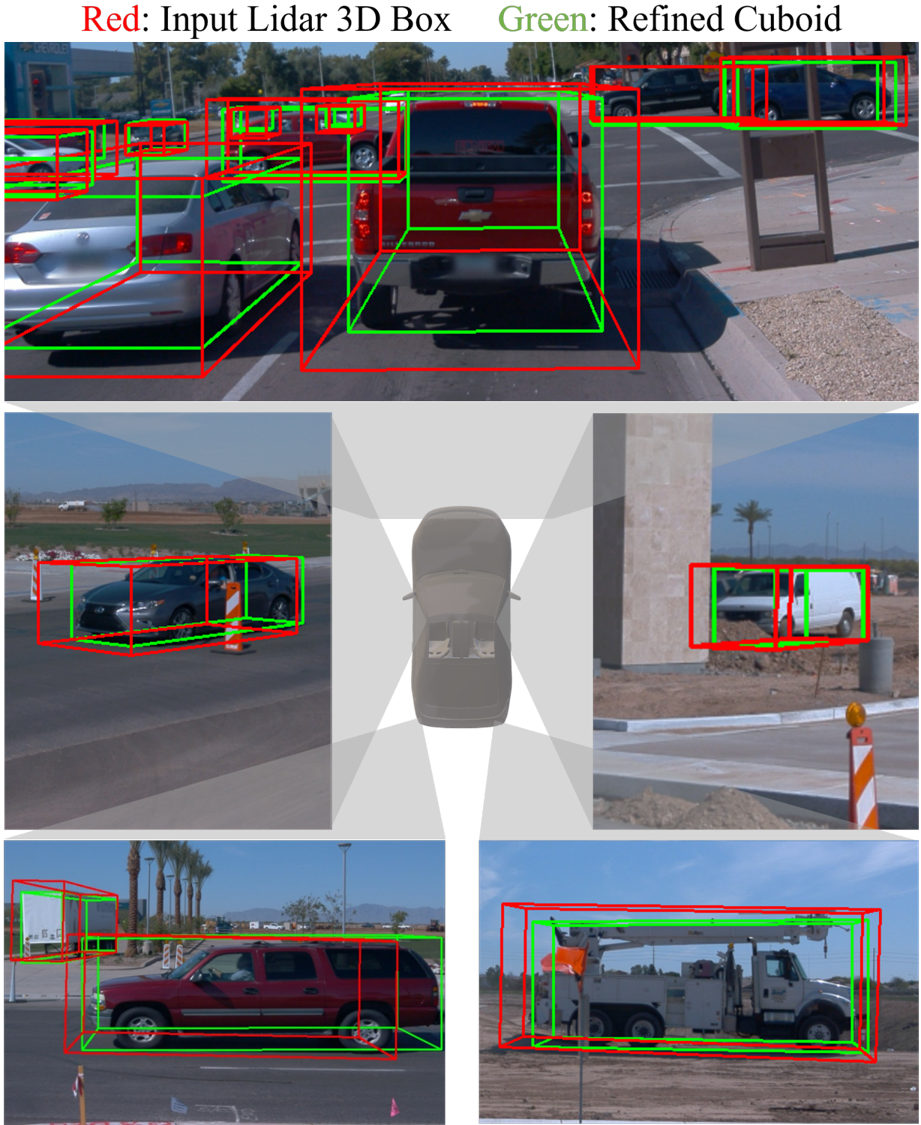}
      }
       \caption{The image projection of the 3D bounding boxes \textcolor{red}{before} and \textcolor{green}{after} refinement, where we show different camera views from the Waymo validation set. Our network can deal well with dislocation, tailing, and different viewpoints and generate 3D cuboids that align well with 2D areas.}
       \label{fig:llus}

    \end{figure}

    With the development of autonomous driving and robotics, the requirements of perception tasks on labeled data become more detailed and rigorous. Early in 2018, autonomous driving practitioners began to employ laborers to label image-based 3D boxes manually\cite{scaleai} as compensation for lidar-based 3D annotations. Such a 3D box, also named cuboid, can be regarded as a 2D bounding box plus position and orientation, which not only provides essential image-level object understanding for redundant perception in compensating for a single LiDAR modality but also can be more in line with the characteristics of monocular 3D tasks and enhance the training of monocular 3D tasks such as monocular 3D detection and depth prediction.

    However, labeling accurate cuboids on visual data is heavily labor-consuming due to the degree of freedom (DoF) complexity and 3D information loss. Currently, there are two main ways to obtain the 3D cuboid annotations. The first method tries reconstructing the object-centric point cloud from sequential image input\cite{ahmadyan2021objectron} and then using the point clouds to guide the annotation. However, the static assumption in reconstruction-based approaches limits its application for dynamic objects in large-scale self-driving scenarios. An alternative approach is rendering differentiable shape models for each object from single depth input \cite{Zakharov2020Autolabeling3O}, but it cannot guarantee temporal consistency. In summary, both reconstruction and rendering methods focus on dense pixel-level shape recovery, which requires much more computation than pure sparse 3D bounding box representation.
    
\begin{figure}[h]
      \centering
      \fbox{
      \includegraphics[width=0.92\linewidth]{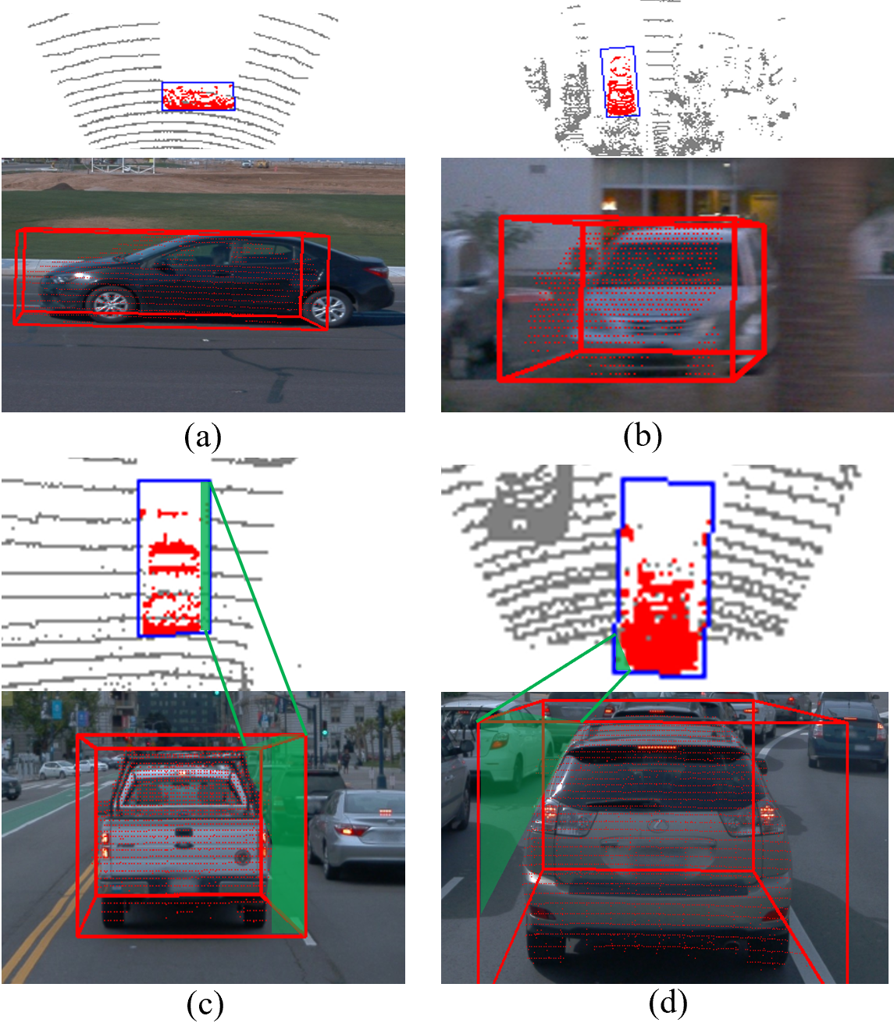}
      }
       \caption{Examples where Lidar boxes cannot be used directly as cuboids. (a),(b): Significant deviation of rear-view images. (c): Misalignment brought by annotation inaccuracy. (d): Gaps because of vehicle shapes. We project the \textcolor{red}{Lidar points} to 2D images to show the position of Lidar objects, and use \textcolor{green}{green} to show the corresponding margins in 2D and 3D.}
        \vspace{-1em}
       \label{fig:reason}
    \end{figure}

    The second widely adopted way is labeling 3D bounding boxes on the Lidar point cloud \cite{Sun2020ScalabilityIP, Caesar2020NuScenesAM}, and then projecting the 3D boxes onto the image as the supervision signal for image-based 3D detection network. However, there are many challenges to obtaining accurate image cuboid annotation from Lidar 3D boxes, such as sensor timestamp synchronization and perfect extrinsic calibration for the whole image area. Even the Waymo dataset team \cite{Sun2020ScalabilityIP}, who made great efforts to overcome the above engineering difficulties, still suffered from several insuperable factors that affect the projection accuracy. For instance, the ``rolling-shutter'' effects will cause high relative speed will cause significant deviation (Fig. \ref{fig:reason}.a, Fig. \ref{fig:reason}.b, ), especially for side view cameras. Even trivial annotation inaccuracy on the 3D Lidar space leads to non-trivial projection misalignment (we highlight the gap by green in Fig. \ref{fig:reason}.c for better illustration). 
    
    Most importantly and most commonly, the vehicle is not a perfect cuboid shape (marked as green triangular corner for the left tail-lamp area in Fig. \ref{fig:reason}.d). That is why most 3D box projections look enlarged compared to the actual image area, even using the perfect 3D bounding box annotation. All the above reasons jointly conclude that directly projecting the 3D bounding box is insufficient for precise image-based learning tasks. As a result, both Waymo\cite{Sun2020ScalabilityIP} and NuImage\cite{Caesar2020NuScenesAM} employ additional annotators to label image 2D bounding box for complimentary. This ``redundant'' annotation effort on both 3D and 2D space motivates this research. We argue that a properly trained neural network can refine the projection misalignment caused by various reasons with very few 2D bounding box labels as guidance. You only label once on the 3D Lidar point cloud, while image cuboids and 2D annotations are obtained automatically.

     Our proposed network optimizes minimum parameters in a 3D box based on Lidar annotation and aligns it with 2D images in an end-to-end way, where only the parameters that do not affect the potential collisions and egocentric distance\cite{Deng2021Revisiting3O} between the target and the ego vehicle are refined. Specifically, we keep the position of the nearest face of the target 3D box unchanged and only slightly refine the peripheral vertexes by optimizing the corresponding dimension as Fig. \ref{fig:lshape} illustrated. Our method does not require the cuboid annotation for training but only uses 2D boxes as the guidance signal to ensure that our refined cuboids perfectly fit the actual 2D regions as Fig. \ref{fig:llus} showed. In addition, we jointly train part of 2D labeled images and a large amount of 2D unlabeled images in a semi-supervised manner. We test our approach on the Waymo\cite{Sun2020ScalabilityIP} and NuScenes\cite{Caesar2020NuScenesAM} and compare our method with traditional geometric solvers. The experiments prove that our method applies to various camera settings, such as front view, side view, and rear view, and is helpful to the training and execution of subsequent perceptual tasks. To conclude, our contributions are as follows,

    \begin{itemize}
        \item [1.] Employ the egocentric distance to transform the original ill-posed 3D-2D box alignment problem into a solvable optimization task.
        \item [2.] Propose a training-based framework that outperforms traditional geometric solvers and operates in an end-to-end way.
        \item [3.] Our method is the first to focus on image-level cuboid adaptation. The refinement result can work as the ground-truth value of perception tasks such as monocular 3D detection based on 2D images.
    \end{itemize}

\begin{figure*}
\vspace{1em}
      \centering
      \fbox{
      \includegraphics[width=0.9\linewidth]{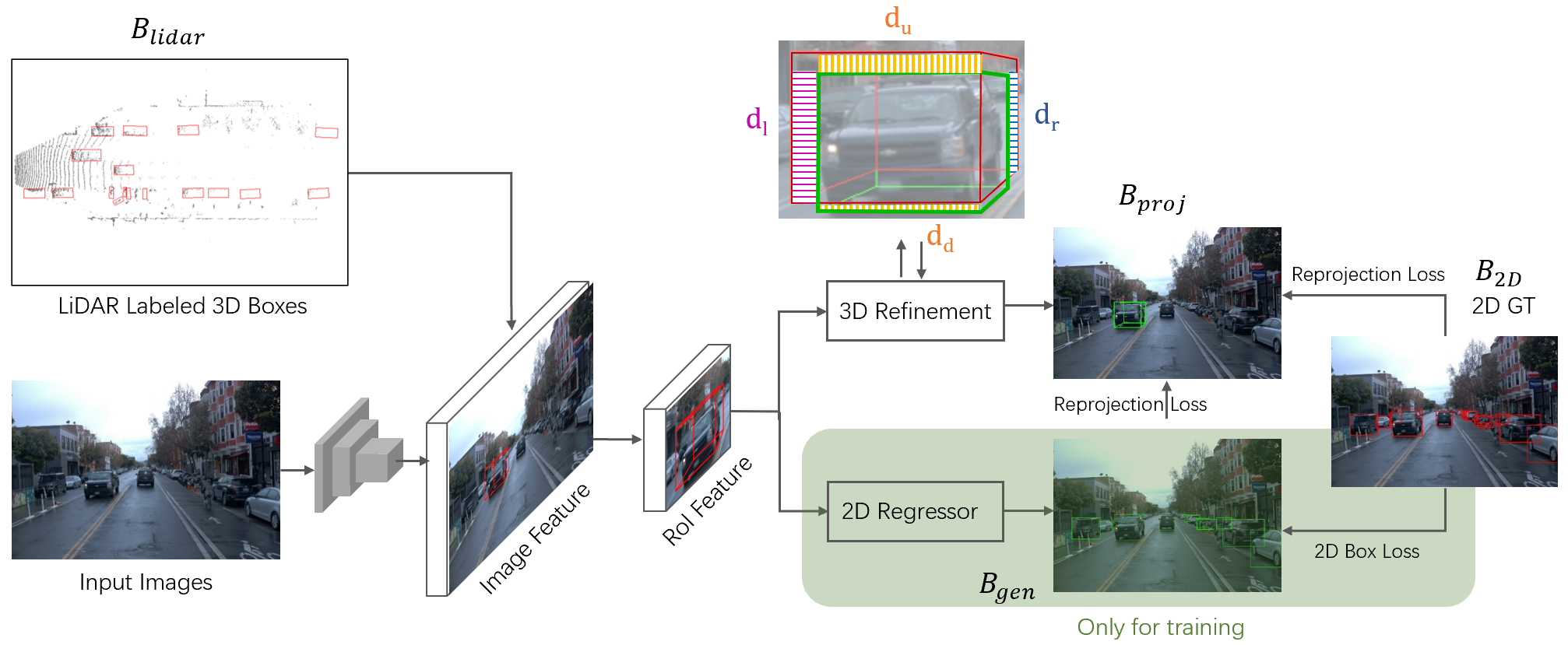}
      }
       \caption{The framework of our network. We use a Faster-RCNN backbone (ResNet50) to extract pyramid feature maps from input 2D images and use 3D Lidar annotations as proposals to crop out corresponding features. The features are then fed to two branches for refined 2D boxes and 3D cuboids. The 2D branch is only used to assist with training and will not participate in the final inference.}
       \label{fig:framework}
\vspace{-2em}
    \end{figure*}

\section{Related Work}
\label{sec:related}
The auxiliary annotation system is an essential part of the closed loop for data processing in autonomous driving. This section briefly introduces the standard automated and semi-automated annotation methods based on Lidar and images.
\vspace{-1em}
\subsection{Lidar-Centric Labeling Methods}
Lidar-Centric labeling methods can further be divided into two kinds. The first is the reconstruction-based method targeting large static scenes, which is usually closely combined with establishing high-precision maps. Early researchers build a complete map of the large scene using the SLAM technology, filter out moving obstacles, and split the whole scenery into small segments for further semantic segmentation and annotation \cite{Bloembergen2021AutomaticLO}. Recently, powered by the BEV-based fusion networks, researchers fuse Lidar point clouds with surrounding images in bird-eye view and integrate semantic segmentation, direction estimation, and some other tasks in a single network to produce a vectorized map\cite{Li2022HDMapNetAO, Zhang2022BEVerseUP}. However, the high requirements for training data and the generalization capability of multi-task models limit the usage of such annotation methods. This inspired some researchers to try simplifying high-precision maps by selecting different kinds of objects as landmarks and creating semantic maps containing only specific semantic objects\cite{Liao2020CoarseToFineVL,Qin2021ALS}. Such semantic maps significantly reduce the previous training work but lead to more accuracy and application limitations.

The other kind of Lidar-based labeling focuses more on moving vehicles. Pioneers use human assistance to confirm the approximate anchor of objects and employ the pre-trained detection and segmentation network to label the objects\cite{Lee2018LeveragingP3,Meng2020WeaklyS3}, or utilizing the consistency between frames to reduce annotation burden\cite{Mei2018SemanticSO}. Researchers have recently tried to minimize the annotators' operation and automatically label objects based on SOTA detection and tracking methods\cite{Qi2021Offboard3O,Yang2021Auto4DLT}. \cite{Qi2021Offboard3O} first leverages a 3D detector to give out the object poses through the whole dataset and then associate objects across different frames using multi-object trackers. Such labeling methods carry out further optimization to improve the accuracy of 3D boxes based on detection networks. However, though succeeding in improving the labeling accuracy, they still rely on the pre-training on well-labeled datasets. \cite{ Najibi2022MotionIU} goes a step further, estimating a 3D scene flow of a sequence of Lidar frames and then accumulating the point clouds belonging to the same object to form 3D amodal bounding boxes as the ground-truth annotations. This method finally eliminates manual annotation but cannot deal with the vacancy and deformation caused by occlusion, rolling shutter, and accumulation error of the optical flow. In addition,
because such labeling methods focus more on Lidar box accuracy, they completely discard 2D information and still need help solving the problems of 3D-2D deviation in cuboid alignment tasks.

\subsection{Image-Centric Labeling Methods}
Image-based labeling methods, lacking spatial information, usually rely on the network to predict the depth map or SDF information and then do the scene-level reconstruction or object-level annotation based on the predicted depth. One typical method is to predict object models using 2D images and supervise the reconstruction result with differentiable shape renderers\cite{Bloembergen2021AutomaticLO}. \cite{Bloembergen2021AutomaticLO} pre-trains a 2D detection network on real data and a shape generation network on CAD datasets, crops out the candidate objects, predicts the SDF of objects, and supervises the whole network using 2D rendering loss. This method avoids complex image depth and shape annotation for real-world data but relies heavily on the CAD dataset used for pre-training. Following Tesla, researchers further try to create ground-truth annotation using Nerf\cite{Mildenhall2020NeRFRS} or improved implicit methods\cite{Zhi2021InPlaceSL,Zhi2021ILabelIN} to avoid such pre-training procedures. However, the shape generation network and the Nerf-based rendering network focus more on pixel-level recovery and are too complex and resource-consuming for the perception task. Other researchers employ 2D-3D consistency, diffusing 2D labels one-on-one onto 3D points\cite{Wang2019LDLS3O, Sautier2022ImagetoLidarSD}, or using 2D-3D constraints to do semantic alignment and assisting the training \cite{Liu2021SemAlignAC, Tian2020UnsupervisedOD}. These methods are usually based on semantic segmentation, which can provide dense supervision signals and reduce the bias caused by incorrect correspondence. Since our cuboid alignment lacks similar constraints, such methods are unsuitable and too resource-consuming. In this paper, we aim to further improve the annotation optimization method, using less manual assistance and lower computational resources while still ensuring as accurate a 2D-3D correspondence as possible.

\section{Problem Definition}
\label{sec:Problem}
Given the 3D Lidar annotation $B_{lidar} = \{x, y, z, h, w, l, r_y\}$, we hope to extract a 3D cuboid $B_{cuboid}$ that can align tightly with 2D objects with only 2D supervision signals $B_{2D} = \{x_{min}, y_{min}, x_{max}, y_{max}\}$. Such a problem, to optimize a 7-D feature with a 4-D loss, is under-constraint and may have an infinite number of solutions. Therefore, our first task is to find more constraints or to keep some variables frozen during the refinement.

We utilize \emph{collisions} introduced by Waymo\cite{Deng2021Revisiting3O} to reduce variables. As is explained in \cite{Deng2021Revisiting3O}, the 3D boxes will be passed to subsequent autonomous driving modules to plan a safe route, and the most critical task is to avoid collisions between vehicles. Such collisions can be determined by the distance between the nearest surface of the target object and the ego-sensor, also known as \emph{egocentric distance}. To avoid errors in the calculation of collisions, we need to keep the \emph{egocentric distance} unchanged during optimization. 

\begin{figure}[h]
     \vspace{1em}
      \centering
      \fbox{
      \includegraphics[width=0.95\linewidth]{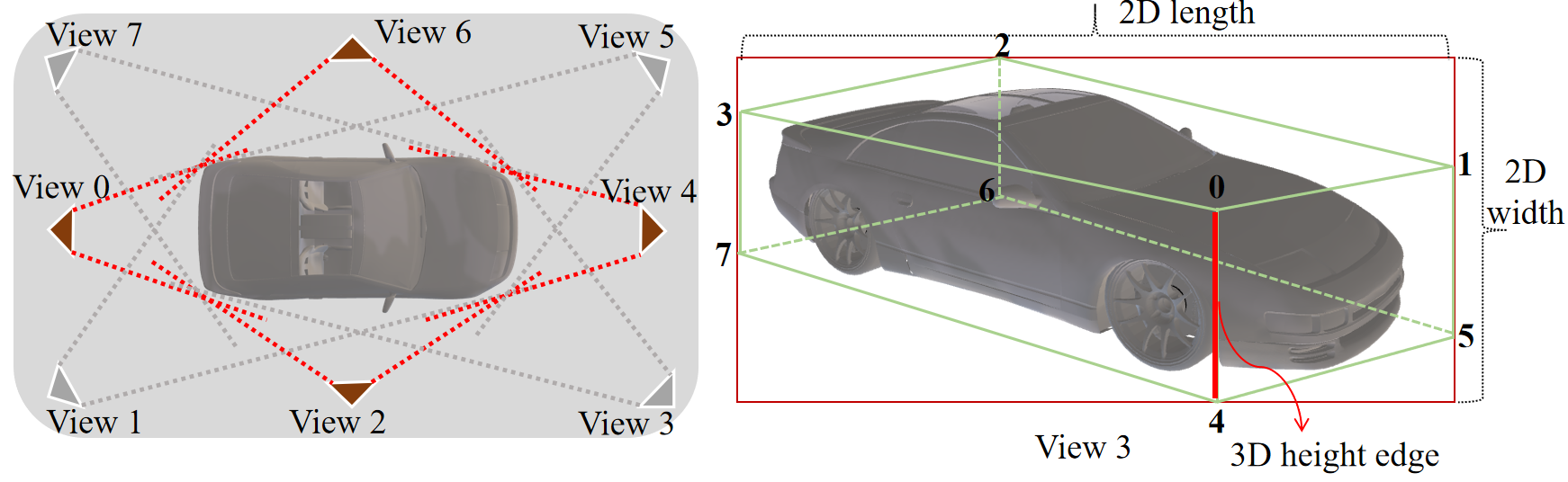}
      }
       \caption{Left: Eight sensor-object relative views. Right: The correspondence between the projected 2D box and 3D view 3. The x-coordinate of the 2D box is determined by 3D corners 3, 7 and corners 1, 5.}
       \label{fig:geometry}
    \end{figure}

\begin{figure}[h]
        \vspace{-1em}
      \centering
      \fbox{
      \includegraphics[width=0.93\linewidth]{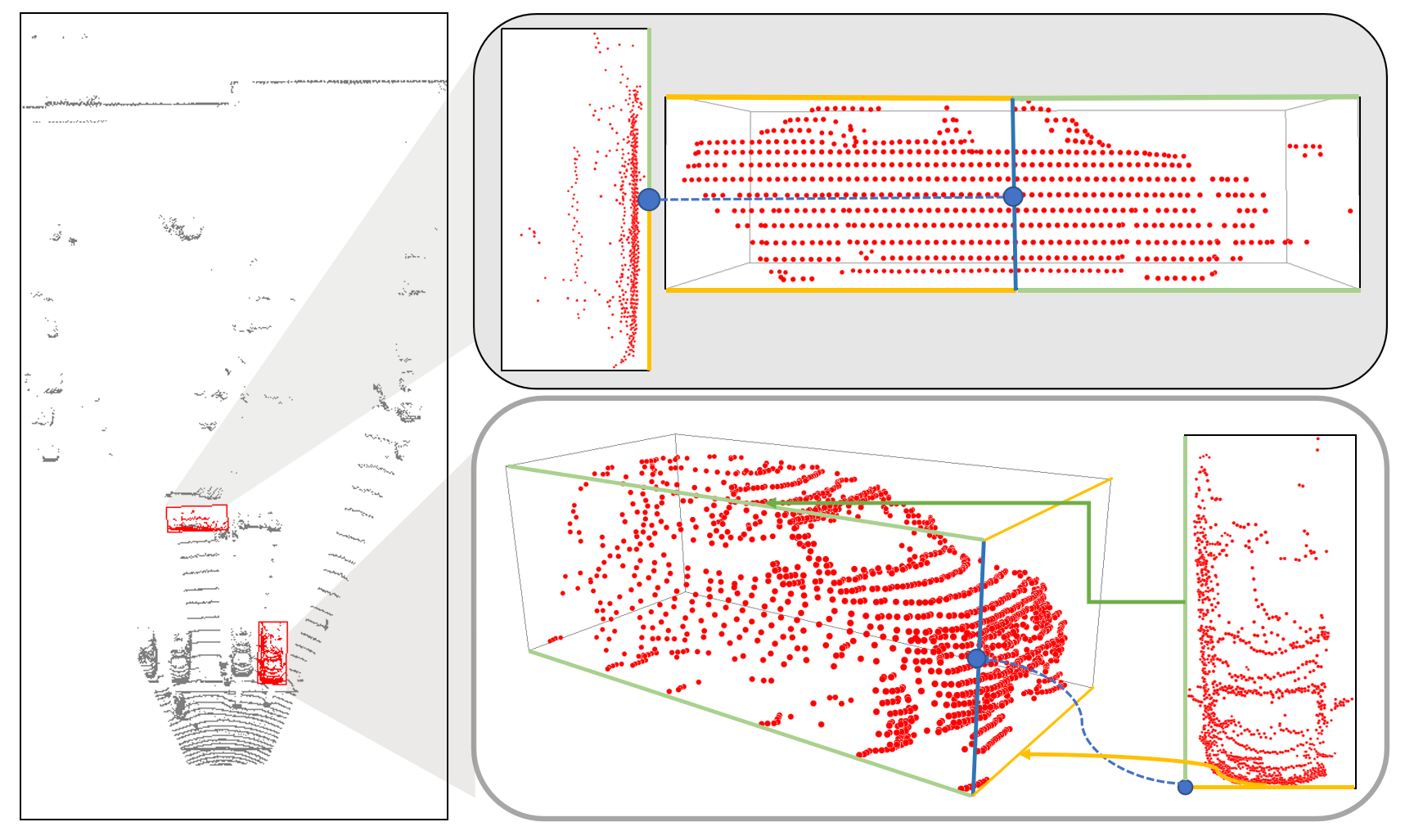}
      }
       \caption{The anchor points on Bev and 3D views. Lower: Rear view case. Upper: Front view case. We use \textcolor{blue}{blue} circle to denote anchor points, and use \textcolor{blue}{blue}, \textcolor{yellow}{yellow} and \textcolor{green}{green} to mark anchor edges. }
       \label{fig:lshape}

    \end{figure}

 We denote the vehicle surface by an anchor point and the plane direction. The plane direction of the nearest vehicle surface corresponds with the original yaw angle $r_y$, and we just need to keep the closest plane changed by selecting and fixing the anchor point. In Fig. \ref{fig:geometry}, we divide the relations between the ego sensor and target object into eight categories based on the viewing angle following\cite{Li2018StereoVS}. In rear-view 1, 3, 5, and 7, we can see two side faces of the target object, and the length of the vehicle's 2D projection is decided by two faces together. For example, in perspective 3 (Fig. \ref{fig:geometry}, right), the x-coordinate of the 2D box is determined by 3D corners 3, 7, and corners 1, 5 together. By changing the horizontal 3D edges corresponding to these corners, we can easily modify the length of the 2D projection. In addition, under the assumption of horizontal ground in autonomous driving scenarios, we can further modify the two endpoints of any 3D height edge to correct the width of the 2D projection box. Therefore, we select the middle of the intersection edge of two side faces, which is the midpoint of the height edge facing the ego-sensor, as the anchor point. Starting from the anchor point, we extend upwards and downwards along the height of the 3D box to obtain two vertical anchor edges and extend left and right along the visible side surfaces for horizontal anchor edges (see Fig.\ref{fig:lshape} (lower)). 

In perspectives 0, 2, 4, and 6, only one side surface is visible, and it determines the length of the vehicle's 2D projection itself. We refer to this situation as a front-view. Since there is only one surface directly affecting the projection result, we select the center of the visible surface as the anchor point and extend it in four directions to obtain the anchor edges (see Fig.\ref{fig:lshape} (upper)). 

It should be emphasized that the selection of anchor points is only related to the original viewpoint, and can be pre-calculated using 2D-3D correspondences\cite{Li2018StereoVS}. We ensure that the egocentric distance remains unchanged by fixing the anchor points and orientations, and optimize the final 2D projection box by modifying the length of the anchor edges. This successfully simplifies our optimization goal into $D_{lidar}$ = $\{d_l, d_r, d_u, d_d\}$ that indicates anchor edge changes.

    \begin{figure*}
    \vspace{1em}
      \centering
      \fbox{
      \includegraphics[width=0.95\linewidth]{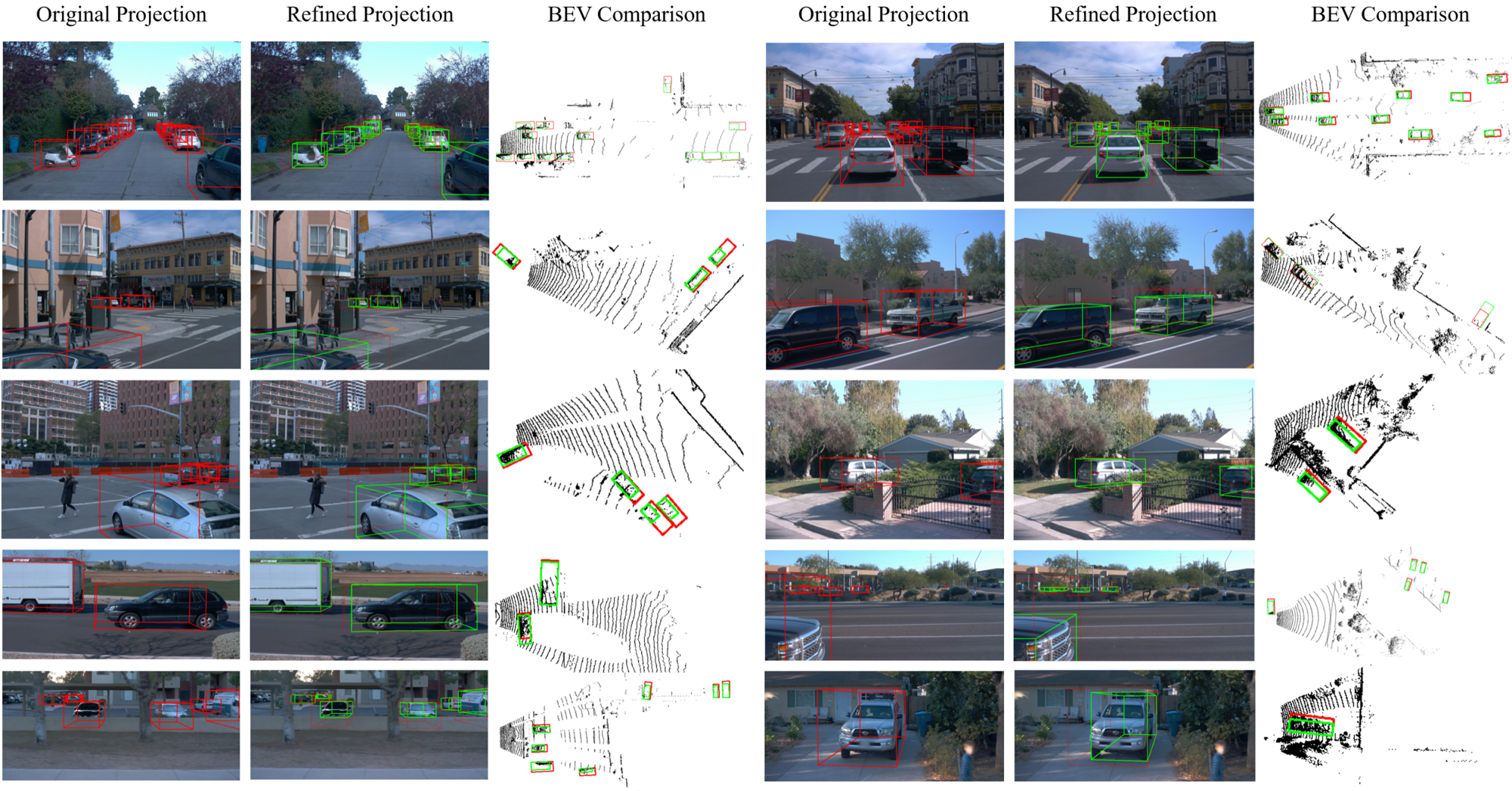}
      }
       \caption{Refinement results on Waymo dataset. We use \textcolor{red}{red} to mark the 3D bounding boxes before refinement and \textcolor{green}{green} to show our optimized results. In order to better show the comparison before and after refinement, we additionally draw the original 3D projection upon our optimized 3D cuboids. We also draw the results in Bev view to show that our refined boxes are still reasonable in geometry. From up to down, we show the results from the front, left, right, rear-left, and rear-right cameras. }
       \label{fig:Waymo}
       \vspace{-2em}
    \end{figure*}

\section{Method}
\label{sec:method}
\subsection{Semi-Supervised Refinement Network}
The cuboid refinement task is completed using a two-stage 2D detection network similar to Faster-RCNN\cite{Ren2015FasterRT}. For a set of input data $(B_{lidar}, I_{RGB})$, we first extract the 2D image features through a simple backbone network (ResNet50 in our experiments) and then project $B_{lidar}$ to 2D to obtain the corresponding features $f_{Lidar}$ for 3D boxes. The projected 2D boxes are regarded as the proposals and, together with the $f_{Lidar}$, fed to a 2D RCNN network for refinement. As mentioned earlier, our network outputs four variables, namely $D_{Lidar} = \{d_l, d_r, d_u, d_d\}$. We use the sigmoid layer to ensure the output $D_{Lidar}$ is within the range of 0-2 to avoid excessive geometric deviation. The original 3D boxes are then refined by $D_{Lidar}$ and projected to 2D again as $B_{proj}$ for loss calculation. 

Since our ultimate goal is to design a helpful labeling tool, the network should be able to conduct self-supervised refinement with as few 2D annotations as possible. Thus, we add an auxiliary 2D branch to the network to reduce or replace the demand for ground-truth data. The simplest auxiliary method is to pre-train a separate 2D detection network and generate 2D pseudo annotations as training signals. However, such training strategies cause a variety of waste. Since our 3D branches are also trained based on 2D images, the two networks can share the backbone, thus reducing the requirement for calculation. 

Meanwhile, different from traditional 2D detection tasks, the input 3D Lidar boxes are the only candidates to be refined, and the RPN stage of popular 2D networks is redundant. Under such consideration, we create an auxiliary branch that shares the same backbone with the 3D network and takes the two-dimensional projection of the 3D box as the proposal to extract 2D features. An RCNN network follows to refine the proposals for the optimized 2D box $B_{gen}$. This design allows us to train the entire network simultaneously when the ground-truth value exists and use the 2D branch to train the 3D results when the 2D ground-truth value is missing.

We emphasize here that we do not specially treat \emph{occlusion} problems but align 3D boxes with 2D annotations. For datasets that label only visible 2D areas, our training results may seem significantly affected by occlusion. However, if the dataset marks the occluded part, our refinement will also consider occlusion during refinement. More results can be found in the experiment section.

\subsection{Loss Function}
During training, our network simultaneously gives out the 2D predicted annotation $B_{gen}$ and the 3D refinement parameter $D_{Lidar}$. The 2D branch is trained following popular 2D detection networks as \cite{Ren2015FasterRT}, and the 2D loss is
\begin{equation}
    E_{2D} = \frac{1}{N_{box}}H(B_{gen} - B_{2D}),
\end{equation} where H denotes the robust Huber loss and $N_{box}$ is the total number of input 3D boxes.

The 3D projection loss is much more complicated than 2D. When an object is too close to the camera, usually the camera can only capture half of the object while the other half is in the blind zones. In such a situation, the object's depth in the camera coordinate system will undergo a negative-to-positive mutation, and the projection result may have an infinite value in the outermost edge, leading to a NaN in the training loss. Therefore, we calculate each edge of the projected box separately and supervise different edges in different situations as
\begin{equation}
    E_{3D} = \frac{1}{N_{box}}\sum_{i=0}^{3} 1_{e^i}H(1_{gt}(B_{proj}^{i} - L_{2D}^{i}) + 1_{!gt}(B_{proj}^{i} - B_{gen}^{i})),
\end{equation}

wherein $1_{e^i}$ is a 0-1 function indicating whether the i-the edge of the projected 2D box is legal. Similarly, $1_{gt}$ and $1_{!gt}$ show whether there is a ground-truth 2D box for training. 

In addition, we set a consistency loss to ensure the consistency of the 3D network and prevent instability during training. For each input box $B_{lidar}$, a random set of variables $D_{aug}$ is generated to augment the input information to $B_{lidar}'$, and to give out the corresponding output $D_{Lidar}'$. If the refinement process of the network is robust enough, we can get $D_{Lidar} = D_{Lidar}' * D_{aug}'$. So during the training, we have

\begin{equation}
E_{con} = \frac{1}{N_{box}}H(D_{Lidar} - D_{Lidar}' * D_{aug}).
\end{equation}

Therefore, the total loss tends to be
\begin{equation}
E_{total} = \lambda_1 1_{gt} E_{2D} + \lambda_2 E_{3D} + \lambda_3 E_{con}.
\end{equation} We use $\lambda_1 = 2.0, \lambda_2 = 3.0, \lambda_3 = 1.0$ in our experiments.

\section{Experiment}
\begin{table*}
\vspace{1em}
  \centering
  \caption{IoU Precision For 2D Boxes Before and After Refinement on Waymo Val Set. }
  \begin{tabular}{c|c|c|c|c|c|c|c}
    \hline
        & Method & Front View & Left Side & Right Side & Left Back & Right Back & Average\\ \hline
        \multirow{3}*{Avg. IoU} & Original & 0.665 & 0.630 & 0.625& 0.619 & 0.590 & 0.639 \\
        & COBYLA & 0.812 & 0.793 & 0.790 & 0.755 & 0.740 & 0.791 \\ 
        & Ours & \bf{0.835} & \bf{0.815} & \bf{0.805} & \bf{0.776} & \bf{0.762} & \bf{0.811} \\ \hline
        \multirow{3}*{Recall(IoU $\ge$ 0.5)} & Original & 82.8\% &76.5\% &75.3\% &73.3\% &69.3\%&78.0\% \\
        & COBYLA & 95.8\% & 93.3\% & 93.4\% & 90.5\% & 88.5\% & 93.5\% \\
        & Ours &\bf{96.3\%} &\bf{94.3\%} &\bf{94.0\%} &\bf{92.1\%} &\bf{91.1\%} &\bf{94.3\%} \\ \hline
        \multirow{3}*{Recall(IoU $\ge$ 0.7)} & Original &48.8\% &40.6\%&39.8\% &37.1\% &29.6\%&42.6\%\\ 
        & COBYLA & 82.6\% & 77.2\% & 77.0\% & 68.5\% & 66.4\% & 77.3\% \\
        & Ours &\bf{85.6\%} &\bf{80.6\%} &\bf{79.1\%} &\bf{73.3\%} &\bf{71.7\%} &\bf{80.4\%} \\ \hline
        \multirow{3}*{Recall(IoU $\ge$ 0.9)} & Original &1.9\% &0.9\% &0.6\% &1.4\% &0.4\%&1.3\% \\ 
        & COBYLA & 30.3\% & 30.6\% & 28.7\% & 21.1\% & 18.5\% & 27.7\% \\
        & Ours &\bf{41.8\%} &\bf{39.3\%} &\bf{35.1\%} &\bf{26.6\%} &\bf{20.9\%} &\bf{36.1\%} \\ \hline
  \end{tabular}
  \label{tab:waymoprecision}
 \vspace{-1em}
\end{table*}      
\label{sec:exp}
\subsection{Experiment Setup}

We use two public data sets, \textbf{Waymo}\cite{Sun2020ScalabilityIP} and \textbf{NuScenes}\cite{Caesar2020NuScenesAM}, to test our cuboid annotation pipeline.

\textbf{Waymo} is an open automatic driving dataset containing over 1,000 video sequences describing different driving scenes. Since there is no public benchmark for our 3D cuboid refinement task on the official test set, we use 798 official training sequences as our training set and 202 official validation sequences as our test set. Only the data corresponding to the VEHICLE tag is used in our experiments. 

\textbf{NuScenes} is also a large-scale autonomous driving dataset with 3D object annotations. However, NuScenes does not provide 2D annotations but uses a separate 2D dataset, NuImages, as a complement. We take the Mask-RCNN\cite{He2020MaskR} model provided by MMDetection\cite{mmdet3D2020} and trained in NuImages as an ``annotator'' to label 2D bounding boxes on NuScenes 3D dataset as ground truth for both training and testing phases.


\textbf{Metric.} Considering there is no ground-truth cuboid annotation on both \textbf{Waymo}\cite{Sun2020ScalabilityIP} and \textbf{NuScenes}\cite{Caesar2020NuScenesAM}, and our refinement does not change the key 3D properties of the labeled 3D bounding boxes as Sect.~\ref{sec:Problem} described, we evaluate the refinement accuracy by the 2D IoU between the cuboid projection and the ground truth 2D boxes, and also the corresponding 2D recall. We also use 3D and BEV IoU to evaluate whether the changed 3D box remains reasonable.

During data preparation, we calculate the IoU between the 3D lidar projection and the 2D annotation and use the Hungarian algorithm to match boxes with IoU more than 0.3. Such matches are used as GT correspondences during training and testing. Unless otherwise specified, only \emph{one-third} of training data is with 2D GT boxes as supervision, and the rest are self-supervised. We train our network for ten epochs from scratch with batch size 16. All the experiments are conducted on a single GTX 1080 Ti GPU.
 \vspace{-1em}
\subsection{Refinement Accuracy}
\subsubsection{Results on the Waymo Val set}
 $\ \ $ To cooperate with five cameras in the Waymo dataset, we pad all images into the same size for joint training and testing. The average IoU and the recall with different thresholds are listed in Table \ref{tab:waymoprecision}. We report the experiment results before and after the refinement for each entry. We also use COBYLA\cite{Powell2007AVO,M1998DirectSA} as a baseline geometric solver as Sec.\ref{sec:cuboid} and list the result in the table. Considering that the refinement difficulty of five cameras varies greatly due to different relative motions, we 
present the evaluation results of the front-view, side-view, and rear-view cameras separately. It can be seen that we achieve an IoU improvement by 0.17 on average and ensure that over $70\%$ of the rear-view boxes maintain a $70\%$ overlap with 2D annotations, which significantly outperform the direct projection counterpart. 

To compare the effectiveness of our refinement more qualitatively, we visualize the projection of the 3D boxes before and after the refinement on both 2D images and BEV maps in Fig. \ref{fig:Waymo}. Even in the most complex rear-view camera scenes, our method can still handle the significant deviation caused by high relative speed and provides reasonable 3D cuboids. It should be noted that our optimization results do not restore the occluded area since the Waymo dataset only labels visible areas.

\subsubsection{Results on the NuScenes Val set}
$\ \ $We repeat the experiment on NuScenes using the same training parameters. After our refinement, the overlap ratio between 3D projection and 2D detection improves a lot (Table \ref{tab:precision}). We believe that this can reflect the robustness of our network. Without changing the network structure, we can refine and re-label 3D datasets with different 2D-3D deviations.

\begin{table}[h]
  \centering
  \caption{IoU Precision on NuScenes Val Set. }
  \begin{tabular}{c|c|c|c|c}
      \hline
    Recall(IoU) & 0.5 & 0.7 & 0.9 & Avg. \\ \hline
      Orig & 90.1\% & 58.4\% & 3.4\% & 0.703 \\
      Refined & 94.6\% & 80.9\% & 42.0 \% & 0.821 \\
      Improvement & +4.5\% & +22.5\% & +38.6\% & +0.119\\ \hline
  \end{tabular}
  \label{tab:precision}
\end{table}

\begin{figure}[h]
  \centering
  \fbox{
  \includegraphics[width=0.93\linewidth]{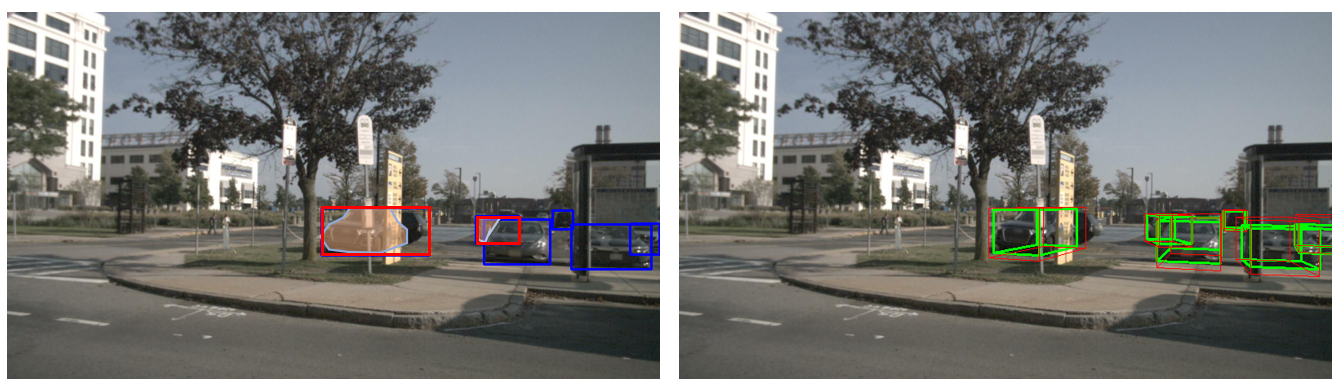}}
   \caption{If the annotators label the occluded 2D parts, our model will also consider occlusion during the refinement.}
   \label{fig:occlusion}
\end{figure}

Unlike the Waymo dataset, which labels only 2D visible areas, the nuScenes data also labels occluded regions in 2D. We show a group of occlusion data randomly selected from nuScenes to validate that our cuboid optimization results are consistent with 2D annotations rather than deliberately processing occlusion in Fig. \ref{fig:occlusion}.
\vspace{-1em}
\vspace{-1em}

\subsection{Consistency Loss for Training}
Our consistency loss hopes to ensure that different 3D boxes corresponding to the same 2D object are regressed to the same output. We use the 2D IoU to evaluate the training results based on small amounts of 2D annotations in Table \ref{tab:consis}. 
\vspace{-1em}
\begin{table}[h]
\centering
  \caption{IoU Difference For Cuboid Projections with and without Consistency Loss}
  \begin{tabular}{c|c|c|c}
    \hline
    GT Num. & Avg. IoU & Recall(0.5) & Recall(0.9) \\\hline
    33.33\%  & +0.02 & +0.0\% & +1.3\%\\\hline
    10\% & +0.04 & +0.2\% & +2.4\%\\\hline
    3.33\% & +0.10 & +0.7\% & +3.1\%\\\hline
  \end{tabular}
  \label{tab:consis}
  \vspace{-1em}
\end{table}  

Our consistency loss further boosts the network's performance with fewer 2D annotations. This is in line with the original intention of our loss design, that is, to ensure the robustness of the network training process and the consistency of the output results and to assist the network in converging when 2D annotations are insufficient.

\subsection{Semi-Supervision Still Works}
To test the accuracy of our semi-supervision strategy in Sect.~\ref{sec:method}, we use the front-view data to do comparative experiments on the Waymo dataset. We randomly select $\frac{1}{3}, \frac{1}{5}, \frac{1}{10}, \frac{1}{30}$ 2D annotations for training respectively.

\begin{table}[h]
  \centering
  \caption{IoU Precision with Different Amount of 2D GT Annotations}
  \begin{tabular}{c|c|c|c|c|c}
    \hline
    2D GT & 100\%  & 33.33\% & 20\% & 10\% & 3.33\%\\ \hline
         Avg. IoU&0.825&0.822 & 0.815 & 0.803 & 0.763 \\ \hline
         Recall(0.5)&95.9\%& 95.7\% &95.9\% & 95.3\% &93.2\%  \\  \hline
        Recall(0.7) &83.9\% & 83.5\% &82.3\% &80.1\% &73.0\%  \\\hline
        Recall(0.9) &37.9\% & 36.0\% &32.5\% & 27.9\% &13.1\%  \\ \hline
  \end{tabular}
  \label{tab:semi}
\end{table}  

From Table \ref{tab:semi}, our auxiliary supervision method can obtain almost equivalent accuracy using 20\% 2D annotations and can still stably optimize the 3D boxes when using only 5000 2D annotations (1/30). This shows that our network can reduce the demand for labeled data as much as possible.

 \section{Analysis of Cuboid's Advantages}
 \label{sec:cuboid}
We use two sets of experiments to prove the assumption that cuboids are more suitable for 2D-based autonomous driving tasks than lidar boxes (Sec.~\ref{sec:intro}). Firstly, the SOTA monocular 3D detection network ImVoxelNet\cite{Rukhovich2021ImVoxelNetIT} realized by mmdetection3D is employed to prove that our cuboid helps in monocular 3D detection tasks. Secondly, we train a Faster-RCNN network on Waymo and prove that our network can provide a more robust and accurate 2D and cuboid annotation than a simple combination of 2D detection, lidar proposals, and geometric solvers.

\subsection{Cuboids Help in Monocular 3D Tasks}
Considering that most monocular 3D detection networks hold that the vehicles conform to the average size and depth while our network cares less about dimensions, we abandon the traditional mAP and use the accuracy of the 2D projection of eight corners (the popular keypoints in 3D tasks\cite{Li2018StereoVS}) to evaluate the network performance. We train and evaluate ImVoxelNet on the Waymo dataset with lidar boxes and cuboids separately, and calculate the keypoint precision on the validation set (after the nms operation). To reduce the impact of the 2D scale, we calculate $\frac{distance^2}{GT area}$ following mscoco\cite{Lin2014MicrosoftCC}, and use the recall as the indicator in Table \ref{tab:corner}. Obviously, cuboids can provide more accurate keypoint results on the premise that cuboids cannot make full use of dimension apriori. This shows that the cuboid can help the network better learn the picture feature and is more suitable for the mono-3D tasks.
 \vspace{-1em}
\begin{table}[h]
  \centering
  \caption{Keypoint Precision on 2D}
  \begin{tabular}{c|c|c|c|c}
      \hline
    Recall(Distance) & 0.01 & 0.05 & 0.1 & 0.5 \\ \hline
      GT Box & 9.5\% & 25.5\% & 35.1\% & 58.7\%    \\
      Cuboid Box & 11.7\% & 27.2\% & 36.7\% & 60.1\% \\
      Improvement & +2.2\% & +1.7\% & +1.6\% & +1.4\% \\ \hline
  \end{tabular}
  \label{tab:corner}
 \vspace{-1em}
\end{table}

\subsection{Cuboids Work better than 2D Methods}
Intuitively, our learned box refinement can be replaced by traditional methods. With manually labeled 2D boxes, we can use traditional optimization solvers to calculate the same ${d_l, d_r, d_u, d_d}$. Considering this substitutability, we designed two experiments to verify the advantages of our network.

The first experiment is to prove that our method can provide better \emph{2D annotations} than 2D detection. We train a Faster-RCNN model from scratch on the Waymo dataset and use the lidar boxes as ROIs during the inference. Similar to Sec. ~\ref{sec:exp}, we use 2D IoU to evaluate the optimization accuracy in Table \ref{tab:fasterrcnn}. This experiment proves that our method can be used as a more accurate 2D annotator to carry out 2D annotations with the help of lidar boxes and reduce the workload of 2D labeling in autonomous driving.
\vspace{-1em}
\begin{table}[h]
  \centering
  \caption{2D Precision of Faster-RCNN Methods(Avg. IoU)}
  \begin{tabular}{c|c|c|c|c|c}
      \hline
    View & Front & Left & Right & Rear-L & Rear-R \\ \hline
    Faster-RCNN & 0.712 & 0.685 & 0.677 & 0.662 & 0.643 \\ 
     Cuboid & 0.835 & 0.815 & 0.805& 0.776 & 0.762 \\ \hline
      Improvement & +0.123 & +0.130 & +0.128 & +0.114 & +0.119 \\ \hline
  \end{tabular}
  \label{tab:fasterrcnn}
 \vspace{-1em}
\end{table}

In addition, we hope to prove that our method can provide a tighter cuboid annotation than traditional solvers. We replace the proposal layer of the Faster-RCNN model with input 3D lidar boxes and then refine the network on the Waymo dataset. Traditional optimization solvers are then employed to solve $D_{lidar}$ with the network output as the optimization target. Despite 2D-3D consistency, we also evaluate the 3D rationality of both methods using 3D IoU and BEV IoU with input lidar boxes. If the network can provide accurate 2D alignment while maintaining a higher 3D overlap, our refinement results can be proved to retain a reasonable geometric structure in 3D. We choose the Nelder-Mead method\cite{Nelder1965ASM} and the COBYLA method\cite{Powell2007AVO,Powell1998DirectSA} implemented by the SciPy library, set the upper and lower bound to 2 and 0 and iterate for 1000 times. 

    \begin{figure}[h]
      \centering
      \fbox{
      \includegraphics[width=0.87\linewidth]{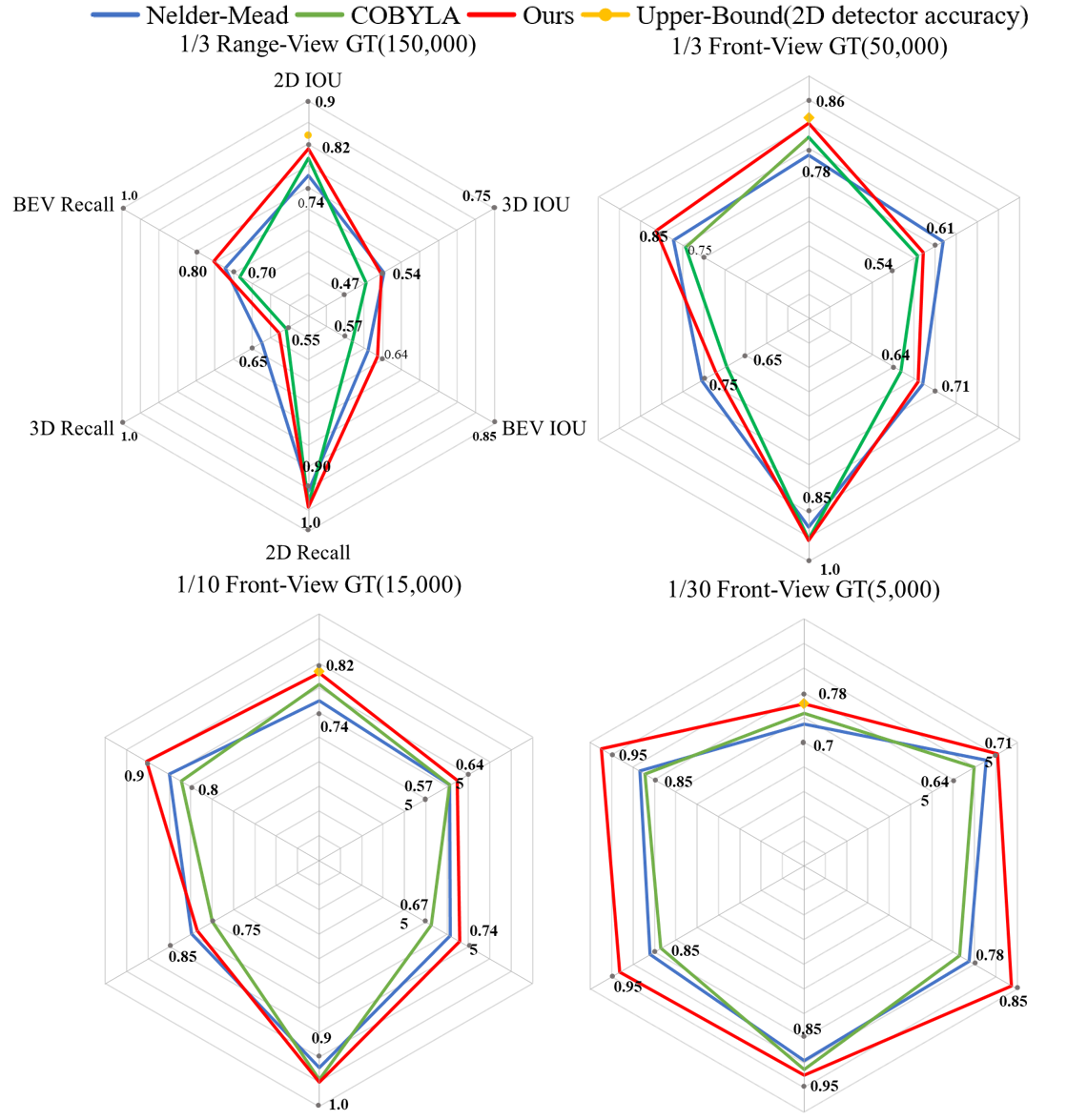}
      }
       \caption{Radar maps showing the difference between our method and optimization solvers. (a): Model trained on the whole Waymo dataset with $\frac{1}{3}$ 2D GT annotations. (b),(c),(d): Model trained on the front-view camera with different numbers of 2D GT.}
       \label{fig:Lidarllus}
        \vspace{-1em}
    \end{figure}

To fully compare the difference between our method and the traditional solvers, we first conduct the standard experiments with $\frac{1}{3}$ 2D annotations on the whole Waymo dataset and illustrate in Fig.~\ref{fig:Lidarllus}.a that our method achieves the overall best performance in both 2D accuracy and 3D IoU retaining. Since the full dataset is too large, we additionally use the front-view images to show that our method can present more and more advantages than the optimization-based methods when the ground truth reduces.

 From Fig. \ref{fig:Lidarllus} and Table \ref{tab:waymoprecision}, our method keeps much higher precision in 2D overlap and leads in BEV overlap, ensuring the best accuracy and avoiding the local minimum as optimization-based methods. Secondly, when the 2D GT data decreases, the 2D gap between the three methods narrows, and our method gradually gains advantages in the 3D overlap. This trend shows that when guidance information is insufficient, our method can obtain the maximum 2D accuracy through the minimum modification of the input box. It also reflects that our method requires fewer 2D annotations than traditional methods.

\section{Conclusions and Discussions}
We propose a semi-supervised label refinement method, which can replace the manual adjustment process and convert a 3D box based on Lidar into a 3D cuboid based on the camera. Our method takes collisions between vehicles as the starting point and uses the egocentric distance of objects to reduce free variables. With a lightweight 2D auxiliary branch and a simple consistency loss, we conduct semi-supervised training with a small number of 2D GT annotations. Experiments show that our method can produce human-level annotations and deal with various 2D-3D deviation errors efficiently and accurately. 

\bibliographystyle{IEEEtran}
\bibliography{egbib}

\begin{thebibliography}{10}
\providecommand{\url}[1]{#1}
\csname url@rmstyle\endcsname
\providecommand{\newblock}{\relax}
\providecommand{\bibinfo}[2]{#2}
\providecommand\BIBentrySTDinterwordspacing{\spaceskip=0pt\relax}
\providecommand\BIBentryALTinterwordstretchfactor{4}
\providecommand\BIBentryALTinterwordspacing{\spaceskip=\fontdimen2\font plus
\BIBentryALTinterwordstretchfactor\fontdimen3\font minus
  \fontdimen4\font\relax}
\providecommand\BIBforeignlanguage[2]{{%
\expandafter\ifx\csname l@#1\endcsname\relax
\typeout{** WARNING: IEEEtran.bst: No hyphenation pattern has been}%
\typeout{** loaded for the language `#1'. Using the pattern for}%
\typeout{** the default language instead.}%
\else
\language=\csname l@#1\endcsname
\fi
#2}}

\bibitem{scaleai}
S.~AI, ``Adding a dimension: Annotating 3d objects with 2d data,''
  \url{https://scale.com/blog/3d-cuboids-annotations}, 2018.

\bibitem{ahmadyan2021objectron}
A.~Ahmadyan, L.~Zhang, A.~Ablavatski, J.~Wei, and M.~Grundmann, ``Objectron: A
  large scale dataset of object-centric videos in the wild with pose
  annotations,'' in \emph{Proceedings of the IEEE/CVF conference on computer
  vision and pattern recognition}, 2021, pp. 7822--7831.

\bibitem{Zakharov2020Autolabeling3O}
S.~Zakharov, W.~Kehl, A.~Bhargava, and A.~Gaidon, ``Autolabeling 3d objects
  with differentiable rendering of sdf shape priors,'' \emph{2020 IEEE/CVF
  Conference on Computer Vision and Pattern Recognition (CVPR)}, pp.
  12\,221--12\,230, 2020.

\bibitem{Sun2020ScalabilityIP}
P.~Sun, H.~Kretzschmar, X.~Dotiwalla, A.~Chouard, V.~Patnaik, P.~Tsui, J.~Guo,
  Y.~Zhou, Y.~Chai, B.~Caine, V.~Vasudevan, W.~Han, J.~Ngiam, H.~Zhao,
  A.~Timofeev, S.~M. Ettinger, M.~Krivokon, A.~Gao, A.~Joshi, Y.~Zhang,
  J.~Shlens, Z.~Chen, and D.~Anguelov, ``Scalability in perception for
  autonomous driving: Waymo open dataset,'' \emph{2020 IEEE/CVF Conference on
  Computer Vision and Pattern Recognition (CVPR)}, pp. 2443--2451, 2020.

\bibitem{Caesar2020NuScenesAM}
H.~Caesar, V.~Bankiti, A.~H. Lang, S.~Vora, V.~E. Liong, Q.~Xu, A.~Krishnan,
  Y.~Pan, G.~Baldan, and O.~Beijbom, ``nuscenes: A multimodal dataset for
  autonomous driving,'' \emph{2020 IEEE/CVF Conference on Computer Vision and
  Pattern Recognition (CVPR)}, pp. 11\,618--11\,628, 2020.

\bibitem{Deng2021Revisiting3O}
B.~Deng, C.~Qi, M.~Najibi, T.~A. Funkhouser, Y.~Zhou, and D.~Anguelov,
  ``Revisiting 3d object detection from an egocentric perspective,'' in
  \emph{Neural Information Processing Systems}, 2021.

\bibitem{Bloembergen2021AutomaticLO}
D.~Bloembergen and C.~Eijgenstein, ``Automatic labelling of urban point clouds
  using data fusion,'' \emph{ArXiv}, vol. abs/2108.13757, 2021.

\bibitem{Li2022HDMapNetAO}
Q.~Li, Y.~Wang, Y.~Wang, and H.~Zhao, ``Hdmapnet: An online hd map construction
  and evaluation framework,'' \emph{2022 International Conference on Robotics
  and Automation (ICRA)}, pp. 4628--4634, 2022.

\bibitem{Zhang2022BEVerseUP}
Y.~Zhang, Z.~H. Zhu, W.~Zheng, J.~Huang, G.~Huang, J.~Zhou, and J.~Lu,
  ``Beverse: Unified perception and prediction in birds-eye-view for
  vision-centric autonomous driving,'' \emph{ArXiv}, vol. abs/2205.09743, 2022.

\bibitem{Liao2020CoarseToFineVL}
Z.~Liao, J.~Shi, X.~Qi, X.~Zhang, W.~Wang, Y.~He, R.~Wei, and X.~Liu,
  ``Coarse-to-fine visual localization using semantic compact map,'' \emph{2020
  3rd International Conference on Control and Robots (ICCR)}, pp. 30--37, 2020.

\bibitem{Qin2021ALS}
T.~Qin, Y.~Zheng, T.~Chen, Y.~Chen, and Q.~Su, ``A light-weight semantic map
  for visual localization towards autonomous driving,'' \emph{2021 IEEE
  International Conference on Robotics and Automation (ICRA)}, pp.
  11\,248--11\,254, 2021.

\bibitem{Lee2018LeveragingP3}
J.~Lee, S.~Walsh, A.~Harakeh, and S.~L. Waslander, ``Leveraging pre-trained 3d
  object detection models for fast ground truth generation,'' \emph{2018 21st
  International Conference on Intelligent Transportation Systems (ITSC)}, pp.
  2504--2510, 2018.

\bibitem{Meng2020WeaklyS3}
Q.~Meng, W.~Wang, T.~Zhou, J.~Shen, L.~V. Gool, and D.~Dai, ``Weakly supervised
  3d object detection from lidar point cloud,'' in \emph{European Conference on
  Computer Vision}, 2020.

\bibitem{Mei2018SemanticSO}
J.~Mei, B.~Gao, D.~Xu, W.~Yao, X.~Zhao, and H.~Zhao, ``Semantic segmentation of
  3d lidar data in dynamic scene using semi-supervised learning,'' \emph{IEEE
  Transactions on Intelligent Transportation Systems}, vol.~21, pp. 2496--2509,
  2018.

\bibitem{Qi2021Offboard3O}
C.~Qi, Y.~Zhou, M.~Najibi, P.~Sun, K.~T. Vo, B.~Deng, and D.~Anguelov,
  ``Offboard 3d object detection from point cloud sequences,'' \emph{2021
  IEEE/CVF Conference on Computer Vision and Pattern Recognition (CVPR)}, pp.
  6130--6140, 2021.

\bibitem{Yang2021Auto4DLT}
B.~Yang, M.~Bai, M.~Liang, W.~Zeng, and R.~Urtasun, ``Auto4d: Learning to label
  4d objects from sequential point clouds,'' \emph{ArXiv}, vol. abs/2101.06586,
  2021.

\bibitem{Najibi2022MotionIU}
M.~Najibi, J.~Ji, Y.~Zhou, C.~Qi, X.~Yan, S.~M. Ettinger, and D.~Anguelov,
  ``Motion inspired unsupervised perception and prediction in autonomous
  driving,'' \emph{ArXiv}, vol. abs/2210.08061, 2022.

\bibitem{Mildenhall2020NeRFRS}
B.~Mildenhall, P.~P. Srinivasan, M.~Tancik, J.~T. Barron, R.~Ramamoorthi, and
  R.~Ng, ``Nerf: Representing scenes as neural radiance fields for view
  synthesis,'' in \emph{ECCV}, 2020.

\bibitem{Zhi2021InPlaceSL}
S.~Zhi, T.~Laidlow, S.~Leutenegger, and A.~J. Davison, ``In-place scene
  labelling and understanding with implicit scene representation,'' \emph{2021
  IEEE/CVF International Conference on Computer Vision (ICCV)}, pp.
  15\,818--15\,827, 2021.

\bibitem{Zhi2021ILabelIN}
S.~Zhi, E.~Sucar, A.~Mouton, I.~Haughton, T.~Laidlow, and A.~J. Davison,
  ``Ilabel: Interactive neural scene labelling,'' 2021.

\bibitem{Wang2019LDLS3O}
B.~H. Wang, W.-L. Chao, Y.~Wang, B.~Hariharan, K.~Q. Weinberger, and M.~E.
  Campbell, ``Ldls: 3-d object segmentation through label diffusion from 2-d
  images,'' \emph{IEEE Robotics and Automation Letters}, vol.~4, pp.
  2902--2909, 2019.

\bibitem{Sautier2022ImagetoLidarSD}
C.~Sautier, G.~Puy, S.~Gidaris, A.~Boulch, A.~Bursuc, and R.~Marlet,
  ``Image-to-lidar self-supervised distillation for autonomous driving data,''
  \emph{2022 IEEE/CVF Conference on Computer Vision and Pattern Recognition
  (CVPR)}, pp. 9881--9891, 2022.

\bibitem{Liu2021SemAlignAC}
Z.~Liu, H.~Tang, S.~Zhu, and S.~Han, ``Semalign: Annotation-free camera-lidar
  calibration with semantic alignment loss,'' \emph{2021 IEEE/RSJ International
  Conference on Intelligent Robots and Systems (IROS)}, pp. 8845--8851, 2021.

\bibitem{Tian2020UnsupervisedOD}
H.~Tian, Y.~Chen, J.~Dai, Z.~Zhang, and X.~Zhu, ``Unsupervised object detection
  with lidar clues,'' \emph{2021 IEEE/CVF Conference on Computer Vision and
  Pattern Recognition (CVPR)}, pp. 5958--5968, 2020.

\bibitem{Li2018StereoVS}
P.~Li, T.~Qin, and S.~Shen, ``Stereo vision-based semantic 3d object and
  ego-motion tracking for autonomous driving,'' in \emph{ECCV}, 2018.

\bibitem{Ren2015FasterRT}
S.~Ren, K.~He, R.~B. Girshick, and J.~Sun, ``Faster r-cnn: Towards real-time
  object detection with region proposal networks,'' \emph{IEEE Transactions on
  Pattern Analysis and Machine Intelligence}, vol.~39, pp. 1137--1149, 2015.

\bibitem{He2020MaskR}
K.~He, G.~Gkioxari, P.~Doll{\'a}r, and R.~B. Girshick, ``Mask r-cnn,''
  \emph{IEEE Transactions on Pattern Analysis and Machine Intelligence},
  vol.~42, pp. 386--397, 2020.

\bibitem{mmdet3D2020}
M.~Contributors, ``{MMDetection3D: OpenMMLab} next-generation platform for
  general {3D} object detection,''
  \url{https://github.com/open-mmlab/mmdetection3d}, 2020.

\bibitem{Powell2007AVO}
M.~J.~D. Powell, ``A view of algorithms for optimization without derivatives
  1,'' 2007.

\bibitem{Rukhovich2021ImVoxelNetIT}
D.~D. Rukhovich, A.~Vorontsova, and A.~Konushin, ``Imvoxelnet: Image to voxels
  projection for monocular and multi-view general-purpose 3d object
  detection,'' \emph{2022 IEEE/CVF Winter Conference on Applications of
  Computer Vision (WACV)}, pp. 1265--1274, 2021.

\bibitem{Lin2014MicrosoftCC}
T.-Y. Lin, M.~Maire, S.~J. Belongie, J.~Hays, P.~Perona, D.~Ramanan,
  P.~Doll{\'a}r, and C.~L. Zitnick, ``Microsoft coco: Common objects in
  context,'' in \emph{European Conference on Computer Vision}, 2014.

\bibitem{Nelder1965ASM}
J.~A. Nelder and R.~Mead, ``A simplex method for function minimization,''
  \emph{Comput. J.}, vol.~7, pp. 308--313, 1965.

\bibitem{Powell1998DirectSA}
M.~J.~D. Powell, ``Direct search algorithms for optimization calculations,''
  \emph{Acta Numerica}, vol.~7, pp. 287 -- 336, 1998.

\end{thebibliography}
\end{document}